\documentclass[a4paper, 10pt, conference]{ieeeconf}      

\IEEEoverridecommandlockouts                              

\usepackage[hidelinks]{hyperref}       
\usepackage{url}            
\usepackage{graphicx} 
\usepackage{amsmath} 
\usepackage{amssymb}  
\usepackage{paralist}
\usepackage{blindtext}
\usepackage{algorithm}
\usepackage{algpseudocode}
\usepackage{multirow}
\let\labelindent\relax 
\usepackage{enumitem}
\usepackage{tablefootnote}
\usepackage{booktabs}
\usepackage{xcolor}
\usepackage{lipsum}
\usepackage{inconsolata} 
\usepackage{tikz}

\title{\LARGE \bf
Enhancing Video-Based Robot Failure Detection Using Task Knowledge
}
\author{Santosh Thoduka$^{1,2}$ \and Sebastian Houben$^{1,2}$ \and Juergen Gall$^{3,4}$ \and Paul G. Pl\"{o}ger$^{2}$
\thanks{This work is supported the Graduate Institute at Hochschule Bonn-Rhein-Sieg, the Deutsche Forschungsgemeinschaft (DFG, German Research Foundation) - GA 1927/4-2 (FOR 2535), and the ERC Consolidator Grant FORHUE (101044724).}
\thanks{$^{1}$Fraunhofer Institute for Intelligent Analysis and Information Systems, Germany
        {\tt\small santosh.thoduka@iais.fraunhofer.de, sebastian.houben@iais.fraunhofer.de}}%
\thanks{$^{2}$Hochschule Bonn-Rhein-Sieg, Germany
        {\tt\small paul.ploeger@h-brs.de}}%
\thanks{$^{3}$University of Bonn, Germany
        {\tt\small gall@iai.uni-bonn.de}}%
\thanks{$^{4}$Lamarr Institute for Machine Learning and Artificial Intelligence, Germany}%
}

\begin{document}

\maketitle
\thispagestyle{empty}
\pagestyle{empty}

\begin{tikzpicture}[remember picture,overlay]
    \node[align=center, text width=\textwidth] at ([yshift=-1cm]current page.north) {%
            \small \textcopyright 2025 IEEE.  Personal use of this material is permitted.  Permission from IEEE must be obtained for all other uses, in any current or future media, including reprinting/republishing this material for advertising or promotional purposes, creating new collective works, for resale or redistribution to servers or lists, or reuse of any copyrighted component of this work in other works. The published version can be found at \url{https://doi.org/10.1109/ECMR65884.2025.11162998}%
  };
\end{tikzpicture}

\begin{abstract}
   Robust robotic task execution hinges on the reliable detection of execution failures in order to trigger safe operation modes, recovery strategies, or task replanning. 
   However, many failure detection methods struggle to provide meaningful performance when applied to a variety of real-world scenarios.
   In this paper, we propose a video-based failure detection approach that uses spatio-temporal knowledge in the form of the actions the robot performs and task-relevant objects within the field of view.
   Both pieces of information are available in most robotic scenarios and can thus be readily obtained.
   We demonstrate the effectiveness of our approach on three datasets that we amend, in part, with additional annotations of the aforementioned task-relevant knowledge.
   In light of the results, we also propose a data augmentation method that improves performance by applying variable frame rates to different parts of the video.
   We observe an improvement from 77.9 to 80.0 in F1 score on the ARMBench dataset without additional computational expense and an additional increase to 81.4 with test-time augmentation.   
   The results emphasize the importance of spatio-temporal information during failure detection and suggest further investigation of suitable heuristics in future implementations.
   Code and annotations are available\footnote{\url{https://sthoduka.github.io/using_task_knowledge/}}.
\end{abstract}

\section{Introduction}
\label{sec:intro}
Vision-based failure and anomaly detection models are used to monitor robot task executions~\cite{inceoglu2021fino,thoduka2024icra}, which allows the robot to take measures to rectify the failure or replan to accomplish the task in a different way~\cite{sliwowski2025condition}.
Such models operate on individual images or video clips that capture the entire duration of the task.
When individual images are used as input, the context of the task and previous observations are often ignored, but failures can be detected as soon as they occur.
For video data, most methods typically sub-sample the video into a fixed number of frames suitable for a video classifier, and, in doing so, encompass more context compared to using individual images.
For some types of failures, it is essential to capture the moment when the failure happens (for example, a dropped object), whereas other types of failures (such as failing to open a drawer) can be detected by simply observing the pre- and post-states of the action.
In the latter case, more contextual frames are necessary, whereas in the former case, the frames that capture the moment when the failure occurs are most important.
This variety in failure types motivates detection strategies that are adapted to the task and expected failure types.
In existing approaches, task-specific knowledge is integrated via hard-coded heuristics, such as cropping frames to a fixed region of interest (ROI)~\cite{inceoglu2021fino,mitash2023armbench}, via natural language descriptions of the tasks for vision-language models (VLMs)~\cite{duan2025aha}, or via learning to predict the robot's actions as an auxiliary task~\cite{thoduka2024icra}.
Several approaches make use of task-relevant multimodal data~\cite{cui2020grasp,inceoglu2021fino,thoduka2024icra}, by combining visual data with tactile, auditory and proprioceptive sensor data.
\begin{figure}[tpb]
   \centering
   \includegraphics[width=0.99\linewidth]{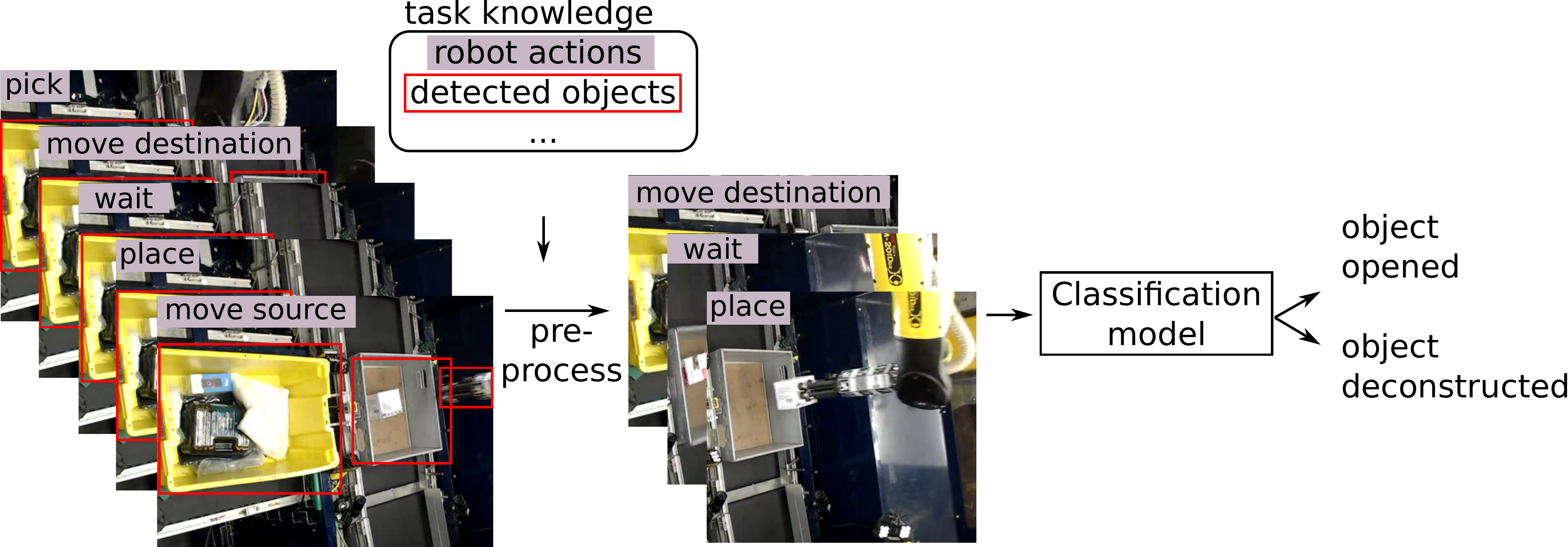}
   \vspace{-4mm}
   \caption{Task knowledge such as the temporal boundaries of the robot's actions and detected task-relevant objects are used in the video pre-processing step to improve failure detection.}
   \label{fig:concept}
    \vspace{-4mm}
\end{figure}

In this paper, we propose to make use of task knowledge to improve performance on the failure detection task, and evaluate our approach on three video-based failure detection datasets.
To this end, we use the temporal boundaries of the robot's actions and the location of task-relevant objects to guide frame selection and pre-processing, shown in Fig.~\ref{fig:concept}, which we extend into a training and test-time data augmentation technique.

The contributions of our paper are:
\begin{inparaenum}[\itshape a)\upshape]
    \item a thorough revision and amendments to the ARMBench~\cite{mitash2023armbench} dataset including fixing incorrect labels, and annotating the temporal boundaries of the robot's actions and bounding boxes for task-relevant objects,
    \item a detailed evaluation of ad-hoc strategies that make use of task knowledge for video frame selection and pre-processing for video-based failure detection on ARMBench and two smaller datasets, and
    \item a data augmentation method for video data that samples frames at variable rates from different parts of the video.
\end{inparaenum}


\section{Related Work}
Vision-based failure and anomaly detection datasets and methods target both manipulation and navigation tasks, and quite often make use of multimodal sensor data.
For learning-based anomaly detection, only nominal samples are used for training and a high anomaly score at inference time indicates a failure.
In contrast, failure detection is formulated as a supervised learning task, with a fixed set of possible failures being considered.
FINO-Net~\cite{inceoglu2021fino} is a multimodal model that detects failures and identifies the failure class during manipulation tasks such as pouring and pushing.
The work by Cui et al.~\cite{cui2020grasp} and Gohil et al.~\cite{gohil2022sensor} both perform visual-tactile fusion to detect grasp failures.
Robot-human handover failures are detected in the work by Thoduka et al.~\cite{thoduka2024icra}, using a combination of visual and proprioceptive data.
Ji et al.~\cite{ji2022proactive} propose a model that fuses 2D images, point cloud data, and the robot's planned path to detect anomalies during outdoor navigation, and Mantegazza et al.~\cite{mantegazza2022sensing} detect hazards during a robot exploration task using visual anomaly detection.
More recently, VLMs have been used to detect and reason about failures during manipulation tasks~\cite{duan2025aha,agia2024unpacking}, or act as action success detectors within a larger task execution framework~\cite{duan2024manipulate,du2023vision,sliwowski2025condition}.
\begin{figure}[tpb]
   \centering
   \includegraphics[width=0.7\linewidth]{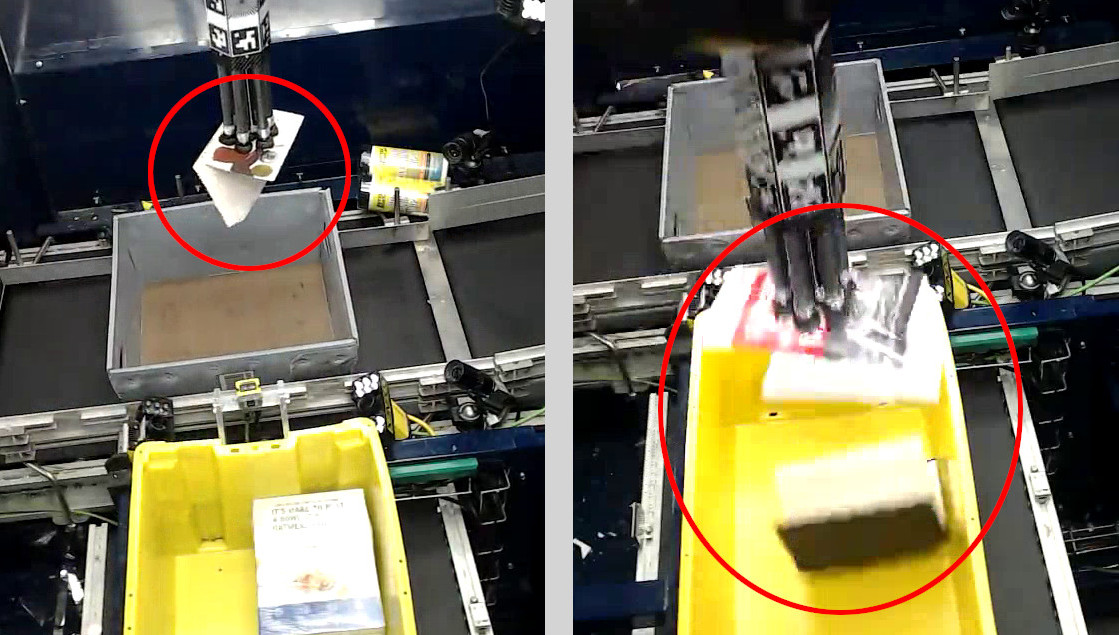}
   \caption{The ARMBench dataset captures \texttt{open} (left) and \texttt{deconstruction} (right) defects during warehouse pick-and-place operations.}
   \label{fig:armbench_ex}
   \vspace{-4mm}
\end{figure}

In addition to multimodal data, some works also make use of task information, the robot's actions, and state information.
For instance, the inputs to VLM-based failure detectors include a natural language description of the task~\cite{duan2025aha,sliwowski2025condition} and an expectation of the nominal state~\cite{agia2024unpacking,duan2024manipulate, sliwowski2025condition}.
The approach by Thoduka et al.~\cite{thoduka2024icra} learns to predict the robot's actions as an auxiliary task during the detection of handover failures, and in~\cite{thoduka2021using}, the known robot motions derived from its internal states are used to detect motion-related anomalies.
Park et al.~\cite{park2018multimodal} use a task progress-based prior for deciding on an anomaly detection threshold during a robot-assisted feeding task.
The robot state and current action being performed have also been used as inputs to networks performing future video prediction~\cite{finn2016unsupervised}, and imitation learning~\cite{jang2022bc}.
In our approach, rather than using the robot's actions as an input, we use their temporal boundaries to select better frames for a video classification model.

For video action recognition models, the common practice for selecting frames is to sample a clip with a fixed stride and duration from the video during training, and to average the softmax scores from multiple clips at test time for predictions~\cite{feichtenhofer2019slowfast,carreira2017quo}.
Some approaches employ more sophisticated strategies for selecting frames, such as motion-guided sampling~\cite{zhi2021mgsampler,xian2024pmi}, learning-based approaches that score the importance of frames~\cite{gowda2021smart,lee2024scalable}, and feature-based similarity measures to reject redundant frames~\cite{yoon2023exploring,shen2025longvu}.
The frame rate of the sampled frames is also an important factor, as shown by Feichtenhofer et al.~\cite{feichtenhofer2019slowfast}.
The authors propose a model with \emph{Slow} and \emph{Fast} pathways, which applies different sampling rates to the same video for each pathway with lateral connections between the pathways.
This is done both at training and inference time to effectively capture both semantic features (such as object categories, colour, etc.) using the  \emph{Slow} pathway, and temporal features (such as fast moving objects) using the \emph{Fast} pathway.
Epstein et al.~\cite{epstein2020oops} use frame rate prediction as a self-supervision signal for video representation learning, applied to unintentional action detection in videos.
Varying the frame rate has been used as a data augmentation method for time-series classification~\cite{leguennec2016data}, speech-related tasks~\cite{afshan2020variable,ravi2022fraug}, and for videos~\cite{zou2023learning}.
More specifically, Zou et al.~\cite{zou2023learning} use \emph{TDrop}, which randomly drops frames and effectively varies the frame rate for those regions.
In our augmentation method, we instead sample larger regions of the video at different frame rates, both based on the underlying temporal boundaries of the robot's actions and randomly, which is most similar to \emph{window warping} applied in~\cite{leguennec2016data}.


\section{Datasets}
There are very few vision-based failure detection datasets, and they are small, typically collected in lab settings, have low variability, contain induced failures rather than natural ones, and do not have large class imbalances~\cite{inceoglu2021fino,thoduka2024icra,sliwowski2025condition,thoduka2021using,wang2019multimodal}.
An exception is the ARMBench Video Defect dataset~\cite{mitash2023armbench}, which we describe in this section, along with the FAILURE~\cite{inceoglu2021fino} and (Im)PerfectPour~\cite{sliwowski2025condition} datasets.
All three datasets contain nominal samples and a fixed number of failure types in the train and test sets, and thus fall into the category of \emph{failure detection}, for which we use supervised learning.
\subsection{ARMBench Video Defect Dataset}

\begin{figure}[tpb]
   \centering
   \includegraphics[width=0.85\linewidth]{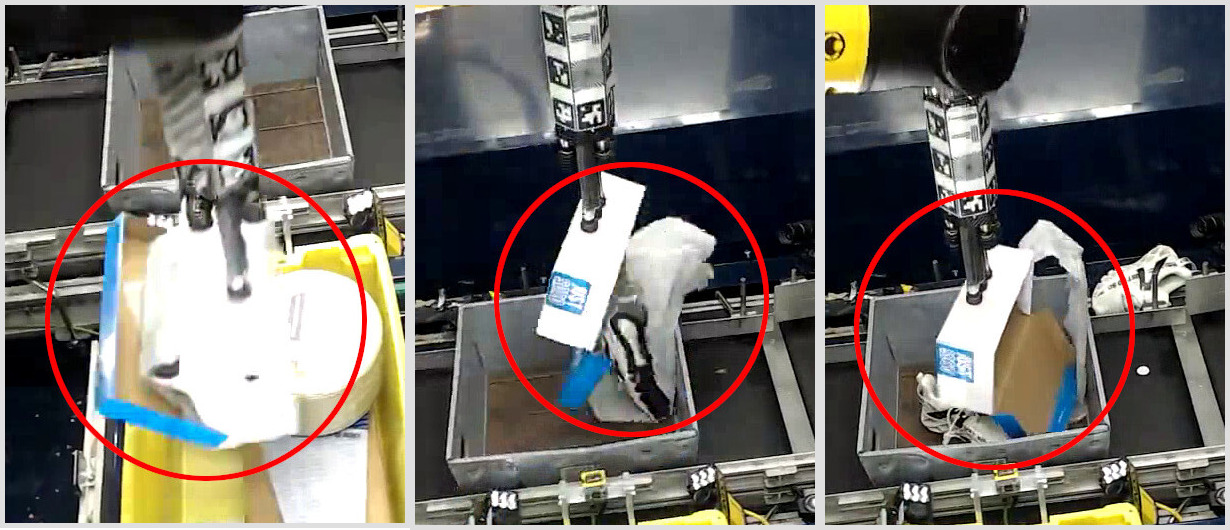}
   \caption{Example of a cascading failure: in the course of a \texttt{deconstruction} failure, an object first opens (left), then deconstructs (center), and finally remains open (right).}
   \label{fig:armbench_open_then_decons}
     \vspace{-4mm}
\end{figure}

\begin{table}[tpb]
  \caption{ARMBench dataset stats}
  \label{tbl:armbench_stats}
  \centering
    \begin{tabular}{l  r  r  r  r}
    \toprule
    \textbf{Split} & \textbf{Nominal} & \textbf{Deconstruction} &  \textbf{Open} & \textbf{Total}\\
    \midrule
    Train          &  6,370  &     1,460   &  927  & 8,757\\
    Test          &  10,175  &     630   &  446  & 11,251\\
   \bottomrule
 \end{tabular}
  \vspace{-4mm}
\end{table}

The ARMBench Video Defect dataset~\cite{mitash2023armbench} consists of 20k video clips of a robot performing pick-and-place operations in a warehouse.
Two types of object defects are caused during the pick-and-place action, as shown in Fig.~\ref{fig:armbench_ex}:
\begin{inparaenum}[\itshape a)\upshape]
    \item \texttt{open}: the manipulated object is opened (e.g. a box cover is lifted)
    \item \texttt{deconstruction}: the manipulated object is deconstructed into multiple pieces (e.g. the contents of a box are spilled).
\end{inparaenum}

The characteristics of the dataset pose some challenges for video-based failure detection:
\begin{inparaenum}
    \item \textbf{Duration and resolution}: The duration of the samples ranges from 2 -- 74 seconds, and the video frames (with a resolution of 1280 x 560) capture two distinct regions where the robot performs its actions (i.e. the regions of the source and destination containers).
    Video classification models typically require clips of a fixed length and square resolution.
    With the large variance in duration, it is challenging to select a suitable frame sampling method, since the coverage of the videos will vary significantly based on their durations.
    \item \textbf{Partial and cascading failures}: As shown in Fig.~\ref{fig:armbench_open_then_decons}, for some objects, the \texttt{deconstruction} failure has three stages: 
    \begin{inparaenum}[\itshape a)\upshape]
        \item the object opens,
        \item the object deconstructs, and
        \item the object remains open after deconstruction.
    \end{inparaenum}
    Since an \texttt{open} object could evolve into a \texttt{deconstruction} failure (or be the result of a \texttt{deconstruction}), it is essential to capture the \emph{right} frames and to have sufficient contextual frames to classify the failure.
    Additionally, for some \texttt{open} failures, a view of the open object is only visible for a short duration, which again stresses the importance of the frame selection strategy.
\end{inparaenum}

\begin{figure}[tpb]
   \centering
    \includegraphics[width=0.85\linewidth]{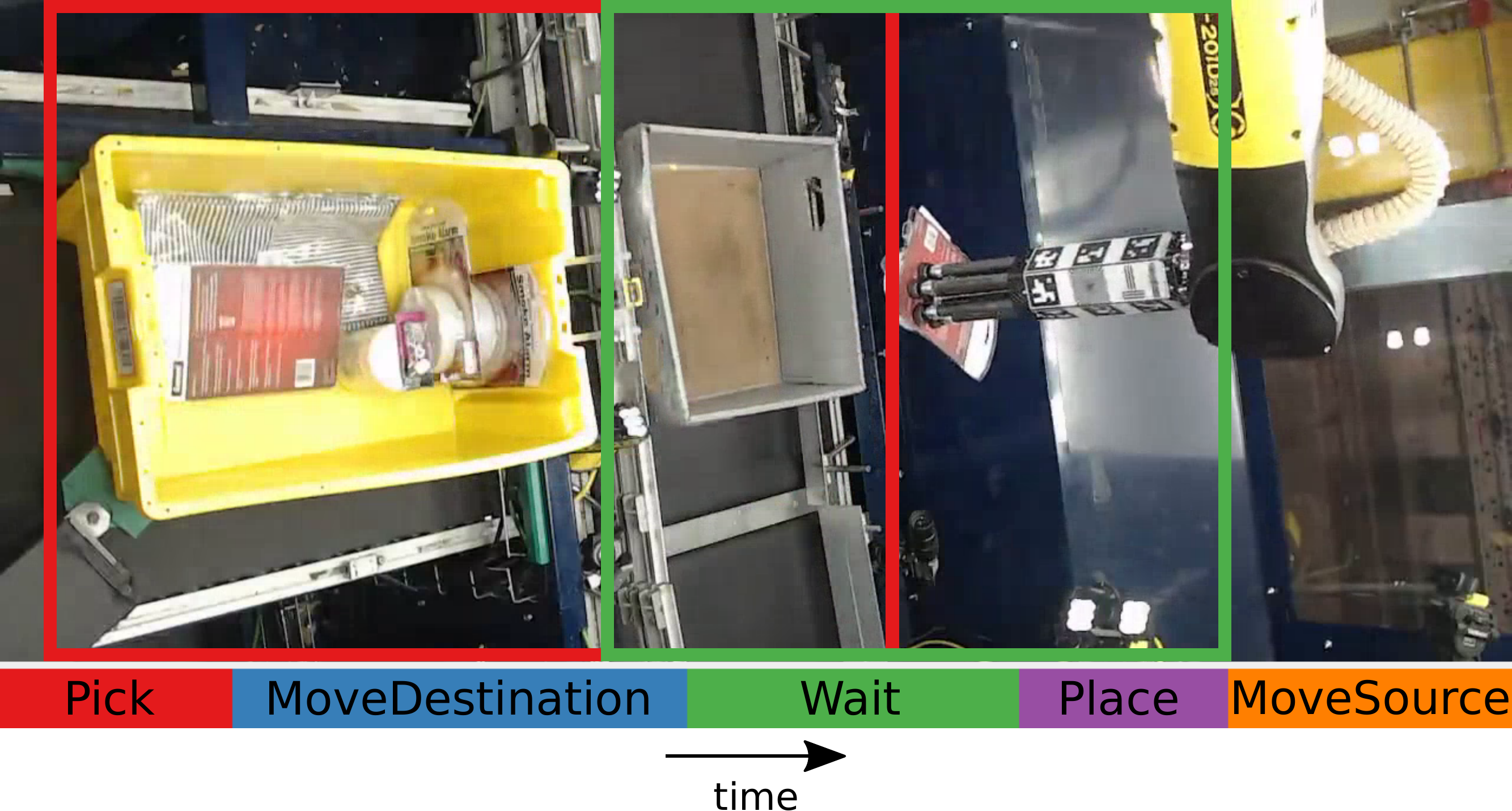}
    \vspace{-4mm}
    \caption{The pick-and-place task in the ARMBench dataset is segmented into the actions shown at the bottom. The current frame shows the \texttt{Wait} action. The red box on the left indicates the cropped region during the \texttt{Pick} and \texttt{MoveDestination} actions, and the green box indicates the cropped region for the remaining actions.}
   \label{fig:armbench_regions}
   \vspace{-4mm}
\end{figure}

We observed that approximately 1100 samples are mislabeled in the original dataset, which we correct.
The data distribution after re-annotation is shown in Table~\ref{tbl:armbench_stats}.
We keep the same train-test split as in the original dataset~\cite{mitash2023armbench}.
We annotate the temporal boundaries of the robot's actions and the bounding boxes of relevant objects in the scene.
In a prototypical execution of the task, the robot performs the following actions in sequence: \(A =\) \{\texttt{Pick}, \texttt{MoveDestination}, \texttt{Wait}, \texttt{Place}, \texttt{MoveSource}\}, as shown in Fig.~\ref{fig:armbench_regions}.
In practice, the temporal boundaries of the actions are known to the robot; here, we label the temporal boundaries using a trained MS-TCN~\cite{farha2019ms} model\footnote{We train the model using 500 manually annotated samples.}.
We also extract bounding boxes for the end-effector, and source and destination containers after training a YOLO-based object detector~\cite{yolov8_ultralytics}.
For videos that contain failures, we annotate the timestamp at which the failure is first visible, with two annotations for \texttt{deconstruction} samples in which the object remains open for a noticeable duration before deconstructing.
The original dataset also contained the failure timestamp annotations, but were found to be largely incorrect.

\subsection{FAILURE Dataset}
The FAILURE dataset~\cite{inceoglu2021fino} consists of 229 clips of successful and failed executions of a robot performing five manipulation tasks on a tabletop.
We annotate the temporal boundaries of the three actions that the robot performs for each task, namely \(A =\) \{\texttt{Approach}, \texttt{Act}, \texttt{Retract}\}, where \texttt{Act} varies based on the task being performed.

\subsection{(Im)PerfectPour Dataset}
The (Im)PerfectPour~\cite{sliwowski2025condition} dataset was developed using a teleoperated robot that prepares drinks.
The tasks include picking, pouring, placing, and wiping, and the dataset contains 1096 samples generated from 544 demonstrations, with failures such as spills, and missing and fallen objects.
Each task is already annotated with temporal boundaries of the actions (referred to as the pre-, core, and post-segments, similar to the set of actions \(A\) of the FAILURE dataset).

\section{Method}
\begin{figure}[tpb]
   \centering
    \includegraphics[width=0.85\linewidth]{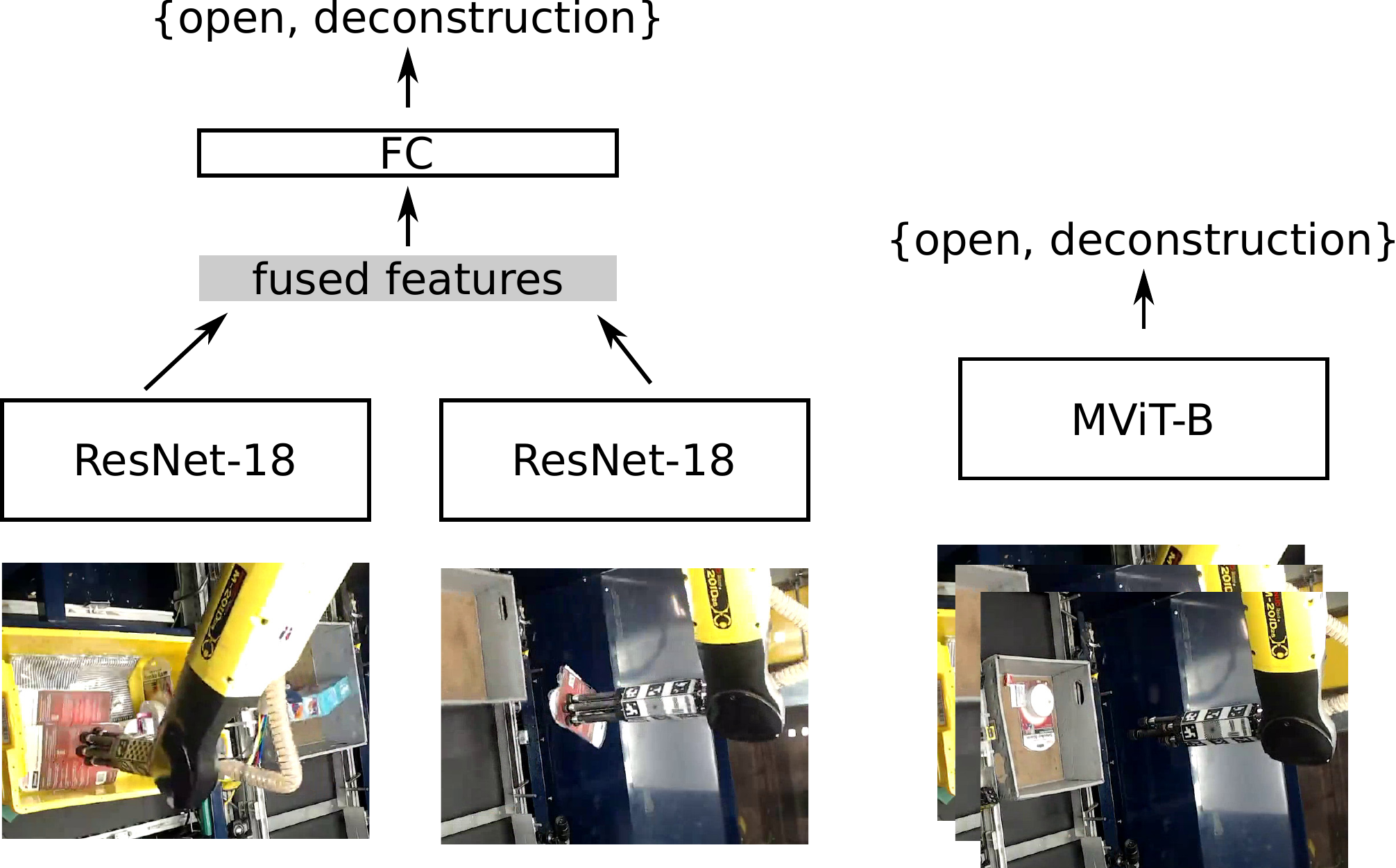}
   \caption{The image pair model (left) predicts the outcome given frames from the pre- and post-states of an action, and the video classification model (right) predicts the outcome given a sequence of frames from the full task or part of the task.}
   \label{fig:models}
    \vspace{-4mm}
\end{figure}

The robot performs a high-level task that is composed of a sequence of \(N\) actions \(A = (a_n)_{n=1}^{N}\), which is recorded by a camera as a sequence of RGB frames, with each action represented by \(l_n\) frames: \((x_i^{a_n})_{i=1}^{l_n}\).
The complete task consists of the frames \(X_{A} = (x_1^{a_1}, \ldots, x_{l_1}^{a_1}, \ldots, x_1^{a_N}, \ldots x_{l_N}^{a_N})\).
Given the frames \(X_{A}\), the task is to perform a binary or multiclass classification to determine the outcome of the task.

\subsection{Models}
In our experiments, we consider the following models.
\subsubsection{Video Classification}
As shown in Fig.~\ref{fig:models}, we use the MViT-B model~\cite{fan2021multiscale} to classify the outcome of the task.
The model expects 32 frames with a resolution of 224 x 224.
\subsubsection{Image Pair Classification}
\label{sec:img_pair_cls}
In nominal executions of an action, the effects of the action are expected to be relatively constant; thus, we hypothesize that the robot can learn to detect failures by observing frames before and after the action is executed.
A similar idea is proposed by Sliwowski and Lee~\cite{sliwowski2025condition}, where they use a transformer-based model to classify whether a frame is in the pre-state, effect-state or neither, and compare the expected and predicted state to detect failures.
As shown in Fig.~\ref{fig:models}, we instead train a ResNet-18 model~\cite{he2016deep} to extract features from a pair of frames, concatenate the features and use a fully-connected (FC) layer to classify the outcome of the task.
For the FAILURE and (Im)PerfectPour datasets, we only classify the \texttt{Act} action, for which we select the first and last frame from the \texttt{Approach} and \texttt{Retract} actions respectively, since they correspond to the pre- and post-states of \texttt{Act}.
Since the actions in ARMBench are more fine-grained, we select the first and last frame for each of the actions in \(A'\), which is defined in Sec.~\ref{sec:action_subset}.
The labels are adjusted based on the state of the failure at the end of each action.
For example, if a \texttt{deconstruction} occurs during the \texttt{Wait} action, and the object is only \texttt{open} during the \texttt{MoveDestination} action, the labels for the \texttt{MoveDestination} and \texttt{Wait} samples from the same video are set to \texttt{open} and \texttt{deconstruction} respectively.
Similarly, if a \texttt{deconstruction} has already occurred prior to the \texttt{Place} action, the label for the \texttt{Place} sample is set to \texttt{open}.
At test time, the predictions for all actions within a task are aggregated, and the final outcome is classified as \texttt{open} only if none of the predicted outcomes are \texttt{deconstruction}.
As an additional baseline, we also report the results using the VGG-RGB model~\cite{inceoglu2021fino}, which consists of a pre-trained VGG-16 model~\cite{simonyan2015very} and convolutional LSTM layers, and expects eight input frames, which we sample uniformly from the full video.

\subsection{Frame Selection and Pre-processing}
\begin{figure}[tpb]
   \centering
    \includegraphics[width=0.99\linewidth]{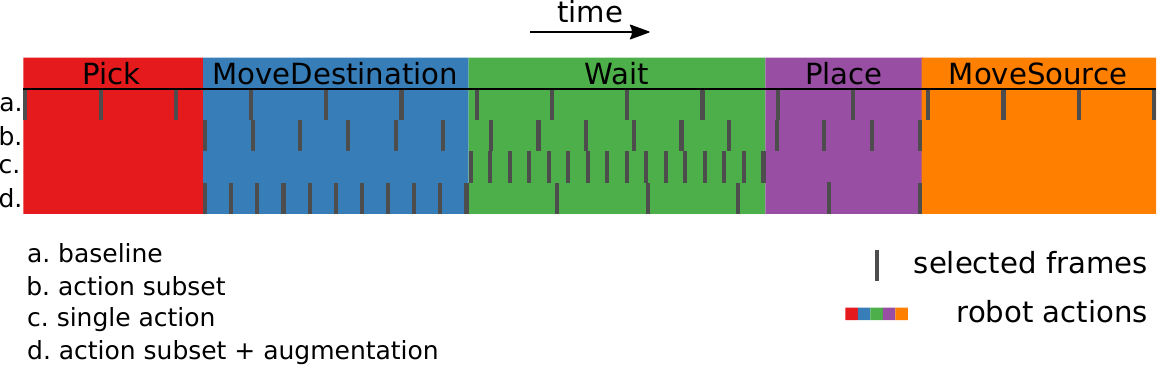}
    \vspace{-4mm}
    \caption{An overview of the different frame selection methods used for training the video classification model.}
   \label{fig:frame_selection}
     \vspace{-4mm}
\end{figure}
To motivate our exploration of frame selection methods, we performed experiments on ARMBench, in which we deliberately selected frames that were likely to lead to better failure detection performance.
For videos with failures, we selected frames in the neighbourhood of the timestamp when the failures are first visible.
For nominal videos, we sampled regions of the video where failures were more likely to occur based on statistics derived from the failure videos.
By training the video classification model on these selected frames, the F1 score improved to 80.0 and 82.1 (without and with region of interest cropping) compared to 77.9 and 77.4 for the baseline with uniform frame sampling under otherwise identical settings.
This indicates the potential that an educated frame selection strategy holds.
These scores can be considered as upper-bounds since we select frames using the ground truth timestamps when failures occurred.

As illustrated in Figs.~\ref{fig:armbench_regions} and~\ref{fig:frame_selection}, for the video classification model trained on ARMBench, we experiment with the following frame selection and pre-processing methods.
The methods rely on two types of task knowledge, namely, the temporal boundaries of each action, and the location of task-relevant objects, which we expect are already known to the robot at no additional expense.
\subsubsection{Baseline}
\label{sec:uniformsampling}
For the baseline, we follow the approach in~\cite{mitash2023armbench}, and sub-sample equidistant frames \(\tilde{x}_{A} \sim \text{Subsample}(X_A, 32)\) from the full video and crop a fixed region of the image before resizing the frames to 224 x 224.
\subsubsection{Action Subset}
\label{sec:action_subset}
We sub-sample equidistant frames from a subset of the actions \(\tilde{x}_{A'} \sim \text{Subsample}(X_{A'}, 32)\), where \(A' = \{\texttt{MoveDestination}, \texttt{Wait}, \texttt{Place}\} \).
We choose these actions since most of the robot-object interaction takes place during these actions, and the failures are most likely to occur during that time.
\subsubsection{Single Action}
\label{sec:actions_separately}
We sub-sample equidistant frames from each action separately \(\tilde{x}_{{a_m}'} \sim \text{Subsample}(X_{{a_m}'}, 32)\), where \({a_m}' \in A'\).
This effectively treats each action within the task as a separate sample, so the model only predicts whether a failure has occurred during that action.
The labels for each action are adjusted as described in Sec.~\ref{sec:img_pair_cls} for image pair classification.
\subsubsection{Variable Frame Rate Data Augmentation}
\label{sec:augmentation}
With the uniform sampling approach for video classification models described in Secs.~\ref{sec:uniformsampling} and~\ref{sec:action_subset}, longer running actions dominate the sampled frames.
While this ensures a good coverage of the entire task, there is a lower representation of the shorter running actions, which may provide valuable context.
In order to balance the requirement for sufficient coverage and context, we propose a train and test-time augmentation method inspired by the SlowFast model~\cite{feichtenhofer2019slowfast}, and similar to the time-series augmentation method proposed by Guennec et al.~\cite{leguennec2016data}.
We sub-sample and concatenate equidistant frames from each action separately \(\tilde{x}_{A'}^{sep} = \{\tilde{x}_{{a_m}'} \sim \text{Subsample}(X_{{a_m}'}, s_m)\}\), where \(s_m\), the number of frames sampled from each action, can vary.
More concretely, for a given sample, we randomly select an action and sample a majority of the 32 frames from that action, and the remaining frames from the other actions.
This results in a clip where the selected action has a high frame-rate and the remaining actions have a low frame-rate.
At inference time, we evaluate by
\begin{inparaenum}[\itshape a)\upshape]
    \item only using the uniform sub-sampling method (i.e. \(\tilde{x}_{A'} \sim \text{Subsample}(X_{A'}, 32)\)), and 
    \item applying test-time augmentation for all actions, and computing the final result based on the mean logits from each augmentation and the uniformly sampled frames.
\end{inparaenum}
We also experiment with sampling at variable frame rates in a non-action-aligned manner, namely by selecting a random position in the video and sampling frames at a higher rate in its neighbourhood, and slower elsewhere.
Unless otherwise specified, we use action-aligned frame sampling.
\subsubsection{Region of Interest Cropping}
We crop the sub-sampled frames based on the action they belong to.
As illustrated in Fig.~\ref{fig:armbench_regions}, for \texttt{Pick} and \texttt{MoveDestination}, we crop the region formed by the union of the bounding boxes (only along the horizontal axis) of the source container and end-effector across all frames within that action, and similarly for \texttt{Wait}, \texttt{Place} and \texttt{MoveSource}, we crop the region formed by the union of the bounding boxes of the destination container and end-effector across all frames within that action.
\subsection{Training and Metrics}
The MViT-B model is initialized with weights trained on the Kinetics-600 dataset~\cite{carreira2017quo}.
Since none of the datasets have validation sets, we train all models for a fixed number of epochs.
For ARMBench, we first train a baseline model for 5 epochs, and use that to initialize all subsequent models, which are trained for a further 10 epochs.
For the FAILURE and (Im)PerfectPour datasets, we train all models for 50 epochs.
At training time, we introduce randomness in the selected frames by varying the starting point for uniform sampling.
All models are trained 3 times, and the mean results are reported.
We report the F1-score for all datasets. For ARMBench, due to the large number of nominal samples in the test split as defined by~\cite{mitash2023armbench}, we also report the recall (Rec.) and false positive rate (FPR) for the failure classes, \texttt{deconstruction} (Decons.) and \texttt{open}.

\section{Results}

\begin{table}[tpb]
  \caption{Results on the ARMBench dataset}
  \label{tbl:armbench_results}
  \centering
    \begin{tabular}{l c r  r  r@{\hspace{2pt}}r r@{\hspace{2pt}}r r}
    \toprule
\textbf{Frames} & \textbf{Aug.} & \textbf{Crop} &  \textbf{F1} & \multicolumn{2}{c}{\textbf{Decons.}} & \multicolumn{2}{c}{\textbf{Open}}\\
\textbf{(Model)}    &    &                              &             & \textbf{Rec.} & \textbf{FPR}          & \textbf{Rec.} & \textbf{FPR} \\
    \midrule
Baseline~\cite{mitash2023armbench}\textsuperscript{*}                          & -     &                              & -    & 79.0 & 3.0 &  69.0 & 23.0 \\
\midrule
Upper bound                & -      &                        & 80.0 & 87.8 & 1.0 &  71.1 & 1.1 \\
(MViT-B)                   & -      & \checkmark             & 82.1 & 88.4 & 0.8 &  73.2 & 1.0 \\
\midrule
                           & -     &                         & 77.9 & 84.8 & 1.1  & 67.9 & 1.1 \\
 Baseline                    & test &                        & 78.6 & 81.9 & 0.6  & 58.6 & 0.4  \\
 (MViT-B)                    & -    & \checkmark             & 77.4 & 83.6 & 1.0  & 69.4 & 1.3 \\
                             & test & \checkmark             & 79.1 & 80.2 & 0.6  & 66.0 & 0.8 \\
\midrule
                              & - &                          & 78.7 & 85.1 & 1.0 & 67.7 & 0.9 \\
 Action subset               & test &                        & 79.6 & 83.9 & 0.8 & 62.9 & 0.5 \\
 (MViT-B)                    &  -   & \checkmark             & 79.6 & 87.3 & 1.1 & 71.2 & 1.1 \\
                             & test & \checkmark             & 81.0 & 86.3 & 0.8 & 68.1 & 0.7 \\
\midrule
                             & train &                      & 78.9 & 85.8 & 1.0 & 68.6 & 1.0 \\
  Action subset             & both &                        & 80.5 & 86.3 & 0.9 & 66.6 & 0.6 \\
  (MViT-B)                  & train & \checkmark            & 80.0 & 87.0 & 1.0 & 69.3 & 1.0 \\
                             & both & \checkmark    & \textbf{81.4} & 87.3 & 0.9 & 67.9 & 0.6 \\
\midrule
Single action                  & -  &                        & 70.8  & 74.6 & 0.8 & 77.9 & 3.5 \\
 (MViT-B)                    &  -   &  \checkmark            & 71.6 & 73.7 & 0.6 & 81.0 & 3.8 \\
\midrule
 Single action  & - &                & 56.8 & 74.0 & 2.2 & 43.7 & 3.6 \\
 (ResNet-18)      & - & \checkmark     & 62.5 & 79.7 & 2.2 & 64.7 & 4.6 \\
\midrule
Eight frames &   -   &                  & 52.3 & 70.1 & 1.9 & 30.8 & 3.5  \\
(VGG-RGB~\cite{inceoglu2021fino}) & - &  \checkmark              & 52.2 & 70.5 & 2.3 & 31.5 & 3.1  \\
\bottomrule
\multicolumn{8}{l}{\footnotesize{\textsuperscript{*}using MViT-B as reported in~\cite{mitash2023armbench}, with incorrect labels}}\\
 \end{tabular}
\end{table}

\begin{table}[tpb]
  \caption{F1 scores for the FAILURE and (Im)PerfectPour datasets}
  \label{tbl:failure_results}
  \centering
    \begin{tabular}{l  c c  c  c  c  c}
    \toprule
    \textbf{Model} &  \textbf{Aug.} & \textbf{FAILURE} & \textbf{(Im)PerfectPour}\\
    \midrule
    FINO-Net-RGB~\cite{inceoglu2021fino} &  - & 85.8\textsuperscript{\(\dag\)} &  74.0\textsuperscript{*} \\
    VGG-RGB~\cite{inceoglu2021fino} &  - & 90.1\textsuperscript{\(\dag\)} &  - \\
    ConditionNET~\cite{sliwowski2025condition} & - & 88.0\textsuperscript{*} &  \textbf{97.0}\textsuperscript{*}  \\
    ResNet-18 & - &\textbf{92.6}~ & 93.9~  \\
    MViT-B & - & 87.0~ & 96.9~ \\
    MViT-B & train & 87.7~ & \textbf{97.1}~ \\
    MViT-B & both & 88.6~ & \textbf{97.3}~ \\
    \bottomrule
    \multicolumn{3}{l}{\footnotesize{as reported in \textsuperscript{*}\cite{sliwowski2025condition} and \textsuperscript{\(\dag\)}\cite{inceoglu2021fino}}} \\
  \end{tabular}
   \vspace{-4mm}
\end{table}

\begin{table}[tpb]
  \caption{Action-aligned  vs random augmentation on ARMBench}
  \label{tbl:aug}
  \centering
    \begin{tabular}{l r r r  r@{\hspace{2pt}}r r@{\hspace{2pt}}r r}
    \toprule
\textbf{Frames} & \textbf{Aug.} & \textbf{Crop} &  \textbf{F1} & \multicolumn{2}{c}{\textbf{Decons.}} & \multicolumn{2}{c}{\textbf{Open}}\\
                   &                                        &                                   &             & \textbf{Rec.} & \textbf{FPR}          & \textbf{Rec.} & \textbf{FPR} \\
\midrule
\multirow{4}{*}{Baseline}          & action &                      & 78.2 & 85.4 & 1.0 & 69.2 & 1.2 \\ 
                              & random &                              & 78.2 & 85.2 & 1.1 & 68.3 & 1.1 \\
                              & action &  \checkmark          & 77.9 & 84.5 & 1.1 & 69.2 & 1.1 \\ 
                              & random &          \checkmark          & 77.5 & 85.2 & 1.3 & 69.6 & 1.2 \\ 
\midrule
\multirow{4}{*}{Action subset}& action &                      & 78.9 & 85.8 & 1.0 & 68.6 & 1.0 \\ 
                             & random &                               & 78.9 & 86.2 & 1.1 & 68.6 & 1.0 \\ 
                             & action & \checkmark            & 80.0 & 87.0 & 1.0 & 69.3 & 1.0 \\ 
                             & random & \checkmark                    & 79.2 & 86.5 & 1.1 & 70.4 & 1.1 \\ 
\bottomrule
 \end{tabular}
  \vspace{-5mm}
\end{table}

The results on ARMBench with the MViT-B, ResNet-18, and VGG-RGB models are presented in Table~\ref{tbl:armbench_results}.
We find that action-based ROI cropping improves performance in all variants except the baseline, particularly improving the recall for \texttt{open} failures.
Limiting the frame selection to a subset of the actions improves performance compared to the baseline as well.
As seen in both Table~\ref{tbl:armbench_results} and~\ref{tbl:failure_results}, using frame rate data augmentation during training improves performance slightly and, as predicted, test-time augmentation further improves performance for all three datasets, although this comes at the expense of additional computation at test time.

In Table~\ref{tbl:armbench_results}, we see that classifying individual actions (\emph{Single action}) lowers performance considerably for the MViT-B model; this is likely because the individual actions do not capture sufficient context necessary to determine what type of failure has occurred.
The high FPR for \texttt{open} in the \emph{Single action} case suggests that a wider coverage of the task is necessary to reliably conclude that a failure is \texttt{deconstruction} rather than \texttt{open}.
We see a similar result for the ResNet-18 model, which has a further drop in performance due to little task context and no motion-related information.
However, the ResNet-18 model performs well on both the FAILURE and (Im)Perfect datasets in Table~\ref{tbl:failure_results}.
The post-\texttt{Act} states for successful and failed executions in both datasets have low intra-class variability, which explains the relatively good performance of the model.
In Table~\ref{tbl:aug}, we compare action-aligned augmentation versus augmenting random parts of the video on ARMBench during training, and we find that there is no significant difference.

Overall, we find that focusing on certain actions and regions based on task knowledge improves performance without additional computational expense, but a sufficient coverage of the task execution is necessary to reliably detect failures.
The data augmentation approach provides a minor improvement as well, with additional computation for test-time augmentation further reducing the FPR.
False negatives in ARMBench include failures that occur in a short time window, and subtle \texttt{open} failures which require a close inspection of the object, suggesting that even denser sampling of frames and closer crops of the object may be required.

\section{Conclusions}
We highlighted the importance of using readily available task information for improving the performance of video-based failure detection models.
The robot's actions and task-relevant objects are used to select and pre-process frames used by the classification model, leading to improved performance without additional computational expense.
The proposed data augmentation approach, which varies the frame rate of sampled frames, leads to additional improvements through train and test-time augmentation.
Future work could explore the use of multimodal data for more intelligent frame selection, though the lack of large-scale multimodal failure datasets remains an open issue.
%
\bibliographystyle{IEEEtran.bst}
\bibliography{IEEEabrv,references.bib}

\end{document}